\documentclass[conference]{IEEEtran}
\IEEEoverridecommandlockouts
\usepackage{cite}
\usepackage{amsmath,amssymb,amsfonts}
\usepackage{algorithmic}
\usepackage{graphicx}
\usepackage{textcomp}
\usepackage{xcolor}
\usepackage{url}
\usepackage{multirow}
\usepackage{multicol}
\usepackage{subcaption}
\def\BibTeX{{\rm B\kern-.05em{\sc i\kern-.025em b}\kern-.08em
    T\kern-.1667em\lower.7ex\hbox{E}\kern-.125emX}}
\begin{document}

\title{Large Multimodal Agents for Accurate Phishing Detection with Enhanced Token Optimization and Cost Reduction}

\author{\IEEEauthorblockN{Fouad Trad}
\IEEEauthorblockA{\textit{Electrical and Computer Engineering} \\
\textit{American University of Beirut}\\
Beirut, Lebanon \\
fat10@mail.aub.edu}
\and
\IEEEauthorblockN{Ali Chehab}
\IEEEauthorblockA{\textit{Electrical and Computer Engineering} \\
\textit{American University of Beirut}\\
Beirut, Lebanon \\
chehab@aub.edu.lb}
}

\maketitle

\begin{abstract}
With the rise of sophisticated phishing attacks, there is a growing need for effective and economical detection solutions. This paper explores the use of large multimodal agents, specifically Gemini 1.5 Flash and GPT-4o mini, to analyze both URLs and webpage screenshots via APIs, thus avoiding the complexities of training and maintaining AI systems. Our findings indicate that integrating these two data types substantially enhances detection performance over using either type alone. However, API usage incurs costs per query that depend on the number of input and output tokens. To address this, we propose a two-tiered agentic approach: initially, one agent assesses the URL, and if inconclusive, a second agent evaluates both the URL and the screenshot. This method not only maintains robust detection performance but also significantly reduces API costs by minimizing unnecessary multi-input queries. Cost analysis shows that with the agentic approach, GPT-4o mini can process about 4.2 times as many websites per \$100 compared to the multimodal approach (107,440 vs. 25,626), and Gemini 1.5 Flash can process about 2.6 times more websites (2,232,142 vs. 862,068). These findings underscore the significant economic benefits of the agentic approach over the multimodal method, providing a viable solution for organizations aiming to leverage advanced AI for phishing detection while controlling expenses.
\end{abstract}


\section{Introduction}
Phishing attacks are a persistent threat in the digital world, challenging traditional detection methods that often struggle to keep pace with the sophisticated and evolving techniques used by cybercriminals. Traditional methods frequently fail to detect attacks that utilize dynamic web content and advanced social engineering tactics \cite{o2021generative, al2020study}. This underscores the need for more effective and scalable detection technologies that are also cost-efficient.

This study introduces the first use of large multimodal models (LMMs) for phishing detection, employing them to analyze both textual and visual website data. Notably, we utilize two models, Gemini 1.5 Flash and GPT-4o mini, accessed via APIs, to determine their effectiveness in detecting phishing websites by analyzing both URLs and webpage screenshots. This approach allows for the use of advanced AI capabilities to analyze websites effectively, requiring only the submission of well-crafted prompts and eliminating the need for training or maintaining AI models.

Our findings indicate a significant increase in detection accuracy when utilizing both types of data compared to models that analyze only one data type. Specifically, the accuracy for our multimodal approach reaches 93\% with Gemini 1.5 Flash and 94\% with GPT-4o mini, while the F1-scores are 92.71\% and 93.68\% respectively. These results are notably higher than those achieved by models analyzing single inputs alone. For instance, when using only URLs, the accuracy was 83\% for Gemini 1.5 Flash and 90\% for GPT-4o mini, with F1-scores of 79.76\% and 89.01\%, respectively. When analyzing images alone, the accuracy dropped to 72.50\% for Gemini 1.5 Flash and 76\% for GPT-4o mini, with F1-scores of 69.61\% and 69.23\%, respectively. These comparisons further emphasize the superior performance of the multimodal approach.

However, the costs associated with API-based multimodal detection can be substantial due to the increased number of tokens sent when including images. To address this, a cost-effective two-tiered agentic approach is proposed. Initially, an AI agent evaluates the URL; if the result is inconclusive, a second agent examines both the URL and the screenshot. This agentic approach maintains high detection performance with an accuracy of 92\% for Gemini 1.5 Flash and 93\% for GPT-4o mini, and F1-scores of 91.30\% and 92.63\%, respectively. Additionally, a major benefit of this staged strategy is that it significantly reduces the number of tokens required by eliminating unnecessary multi-input queries, thereby lowering API costs. Based on a detailed cost analysis, the agentic approach allows for processing approximately 4.2 times more websites compared to the multimodal approach (107,440 vs. 25,626 per \$100) using GPT-4 Turbo, and 2.6 times more websites (2,232,142 vs. 862,068) with Gemini 1.5 Flash. This significant increase in efficiency underscores the economic advantages of the proposed agentic approach, making it an ideal solution for organizations seeking to utilize prompt-engineered LMMs for effective phishing detection while managing expenses, without the need to train or maintain AI models.

In summary, the main contributions of this research are as follows:
\begin{itemize}
    \item Introducing the first use of LMMs for phishing detection, combining URL and image analysis.
    \item Demonstrating that the multimodal approach substantially improves detection accuracy and F1-scores compared to single-modality methods.
    \item Introducing a novel sequential, two-tiered agentic strategy that significantly reduces the costs associated with multimodal detection.
    \item Presenting detailed economic analysis showing the cost-efficiency of the proposed method, supporting its practical application in cybersecurity settings.
\end{itemize}

The remainder of the paper is organized as follows: Section II provides the background and preliminaries essential to our study. Section III reviews related work. Section IV outlines the methodology, while Section V details the experimental setup and presents the results. Section VI discusses the findings from both performance and cost perspectives. Finally, Section VII concludes the paper and suggests directions for future research.

\section{Background and Preliminaries}
\subsection{Large Language Models (LLMs)}
LLMs are advanced AI systems designed to process and generate human-like text. They are built on deep learning architectures, particularly transformer models, which enable them to process vast amounts of textual data and learn complex patterns in language \cite{zhao2023survey}. Notable examples include OpenAI's GPT series \cite{roumeliotis2023chatgpt} and Google's BERT \cite{devlin2018bert}. These models have achieved remarkable performance across various natural language processing tasks, such as text completion, translation, and question-answering, due to their ability to leverage large datasets and sophisticated training techniques \cite{chang2024survey}.

\subsection{Large Multimodal Models (LMMs)}
LMMs extend the capabilities of LLMs by integrating multiple forms of data, such as text, images, and audio. This integration allows LMMs to understand and generate content that is not only linguistically coherent but also contextually rich, reflecting a better understanding of the world \cite{wu2023multimodal}. For instance, models like GPT-4o mini and Gemini 1.5 Flash can analyze both textual descriptions and visual inputs, facilitating applications in diverse fields, including healthcare \cite{mesko2023impact}, security \cite{shi2024shield}, and autonomous driving \cite{cui2024survey}.

The ability to process multimodal data enhances the performance of AI systems in tasks that require a comprehensive understanding of context, such as phishing detection, where both the textual content of a URL and the visual layout of a webpage can provide critical insights into potential threats. 

\subsection{Agentic AI systems}
Agentic AI systems involve multiple autonomous agents collaborating to achieve a common goal. In the context of LLMs and LMMs, multiagent systems can be employed to enhance the models' capabilities and performance \cite{xie2024large}. For example, a team of specialized agents can work together to process different aspects of the input data, such as linguistic features, visual elements, and contextual information. By dividing tasks among agents and allowing them to communicate and coordinate their efforts, multiagent systems can achieve more accurate and efficient results compared to a single monolithic model \cite{sreedhar2024simulating}.
Agentic models in LLMs and LMMs can also enable more flexible and adaptive behavior. Different agents can specialize in specific tasks or domains, allowing the system to handle a wider range of inputs and scenarios. However, the development of agentic systems also introduces challenges related to coordination, communication, and trust among the agents. Ensuring that the agents work together seamlessly and make decisions aligned with the overall system goals is crucial for the successful use of these models in real-world applications.

\subsection{Phishing Detection}
Phishing is a significant cybersecurity threat characterized by deceptive attempts to obtain sensitive information from users, often through fraudulent emails or websites. Traditional phishing detection methods rely on heuristics and rule-based systems, which can be limited in their effectiveness against sophisticated attacks \cite{da2020heuristic}. Recent advancements in machine learning have shown promise in improving phishing detection rates. These models can analyze the linguistic features of phishing attempts, identify anomalies, and better adapt to evolving tactics used by attackers \cite{tang2021survey}. Prompt-engineered and fine-tuned LLMs have also shown potential in phishing URL detection \cite{trad2024prompt}. This work aims to extend these advancements by evaluating the use of LMMs for phishing detection. By leveraging the contextual understanding of language alongside other input types, such as images, these models can enhance the accuracy and efficiency of phishing detection systems, offering a more robust defense against this threat.

\section{Related Work}
Phishing detection has seen substantial evolution over the years, beginning with traditional heuristic and rule-based methods \cite{qabajeh2018recent,da2020heuristic}. Early detection systems primarily relied on identifying known malicious URLs and implementing keyword-based filters to detect phishing websites \cite{ding2019keyword, tan2016phishwho}. While these foundational techniques offered initial solutions, they often suffered from high false-positive rates and were ineffective against sophisticated phishing tactics designed to closely mimic legitimate communications \cite{ding2019keyword}.

The introduction of machine learning marked a significant improvement in phishing detection approaches. Researchers started developing advanced models such as decision trees \cite{machado2017phishing}, support vector machines \cite{altaher2017phishing}, and ensemble methods \cite{ubing2019phishing} that analyze patterns in phishing URLs. By leveraging large datasets, these models improved detection accuracy by identifying features indicative of phishing attempts. With the rise of deep learning, the field progressed further, enabling the analysis of more complex data representations. Convolutional neural networks and recurrent neural networks have been applied to detect phishing through image analysis and natural language processing techniques \cite{alshingiti2023deep, xiao2020cnn, huang2019phishing}. Studies have shown that deep learning models can effectively classify phishing websites, often outperforming traditional ML approaches \cite{yang2021phishing}. However, the requirement for continuous retraining and maintenance makes these models resource-intensive and limits their practical usability.

In recent years, there has been a shift towards integrating LLMs in cybersecurity applications \cite{zhang2024llms, motlagh2024large}. Prompt-engineered LLMs offer an advantage over traditional ML systems by reducing the need for ongoing retraining, as their performance can be dynamically adjusted using prompts, while ongoing maintenance and retraining are managed by the model providers \cite{trad2024prompt, trad2024ensemble}. These models have demonstrated potential across various cybersecurity domains, including phishing detection, where they identify phishing attempts based on textual analysis \cite{ferrag2024generative}.

While previous works have developed multimodal deep learning models for cybersecurity applications \cite{potter2024multimodal, agrafiotis2024advancing}, the exploration of LMMs in this domain is still in its early stages. Prior studies have compared prompt-engineered LMMs and fine-tuned vision transformers (ViTs) in cybersecurity tasks like trigger detection and malware classification \cite{trad2024evaluating}. While prompt-engineered models have shown promising performance in visually evident tasks like trigger detection, their effectiveness diminishes in visually non-evident tasks like malware classification. 

Despite these advancements, research on the application of LMMs specifically for phishing detection remains unexplored. This work aims to bridge this gap by leveraging prompt-engineered LMMs for phishing detection by combining information from both the URL (text) and screenshot (image) of a website. 

Additionally, there is a limited exploration of the economic implications of using LLMs and LMMs, specifically in the context of API costs \cite{wang2024reasoning}. As organizations increasingly adopt AI-driven solutions, optimizing API-related costs is crucial. Our proposed two-tiered agentic approach addresses this challenge by minimizing unnecessary multimodal queries, thereby enhancing cost-efficiency while maintaining robust detection performance.

\section{Methodology}
In this study, we evaluate four distinct approaches for phishing detection: URL-based detection, image-based detection, a multimodal analysis combining both URL and image inputs, and an agentic approach that prioritizes efficiency by introducing a staged evaluation process. Below, we outline each approach, describing how it functions.

\subsection{URL-Based Detection}
The URL-based approach relies solely on the text of the website’s URL to determine if a website is malicious or legitimate. The LMM receives a prompt asking to classify a given website as phishing or legitimate, based on its URL, and returns a response. The prompt used is shown in Figure \ref{fig:approaches} as "Prompt 1". 

\subsection{Image-Based Detection}
The image-based approach focuses on the visual representation of the webpage, typically captured as a screenshot. The LMM analyzes the visual elements—such as logos, layout consistency, and other graphical features—to detect patterns indicative of phishing attempts. The LMM is prompted to classify a website as phishing or legitimate based on its screenshot alone and returns a response. The prompt used is shown in Figure \ref{fig:approaches} as "Prompt 2". 

\subsection{Multimodal Analysis (URL + Image)}
\begin{figure*}[h]
\centerline{\includegraphics[width=\linewidth]{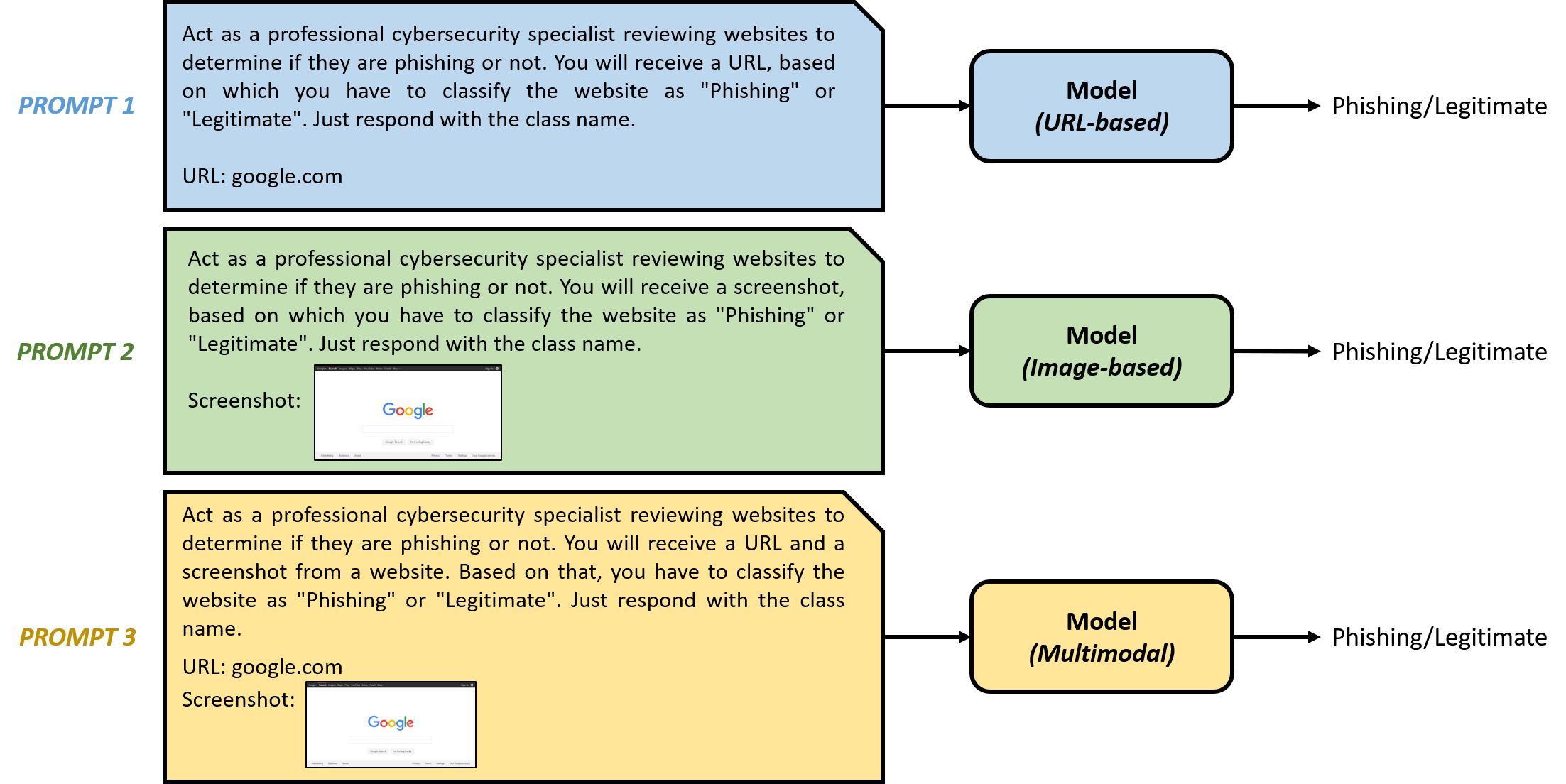}}
\caption{URL-based, Image-based, and Multimodal (URL+image) approaches for Phishing detection}
\label{fig:approaches}
\end{figure*}

The multimodal approach integrates both URL and image data, combining the strengths of both modalities to provide a comprehensive analysis of a potential phishing site. The LMM receives a prompt, as shown in "Prompt 3" in Figure \ref{fig:approaches}, that includes both the URL and a screenshot of the webpage and is asked to classify the website as either phishing or legitimate. The LMM returns a response based on this combined information.

\subsection{Agentic Approach}
\begin{figure*}[h]
\centerline{\includegraphics[width=\linewidth]{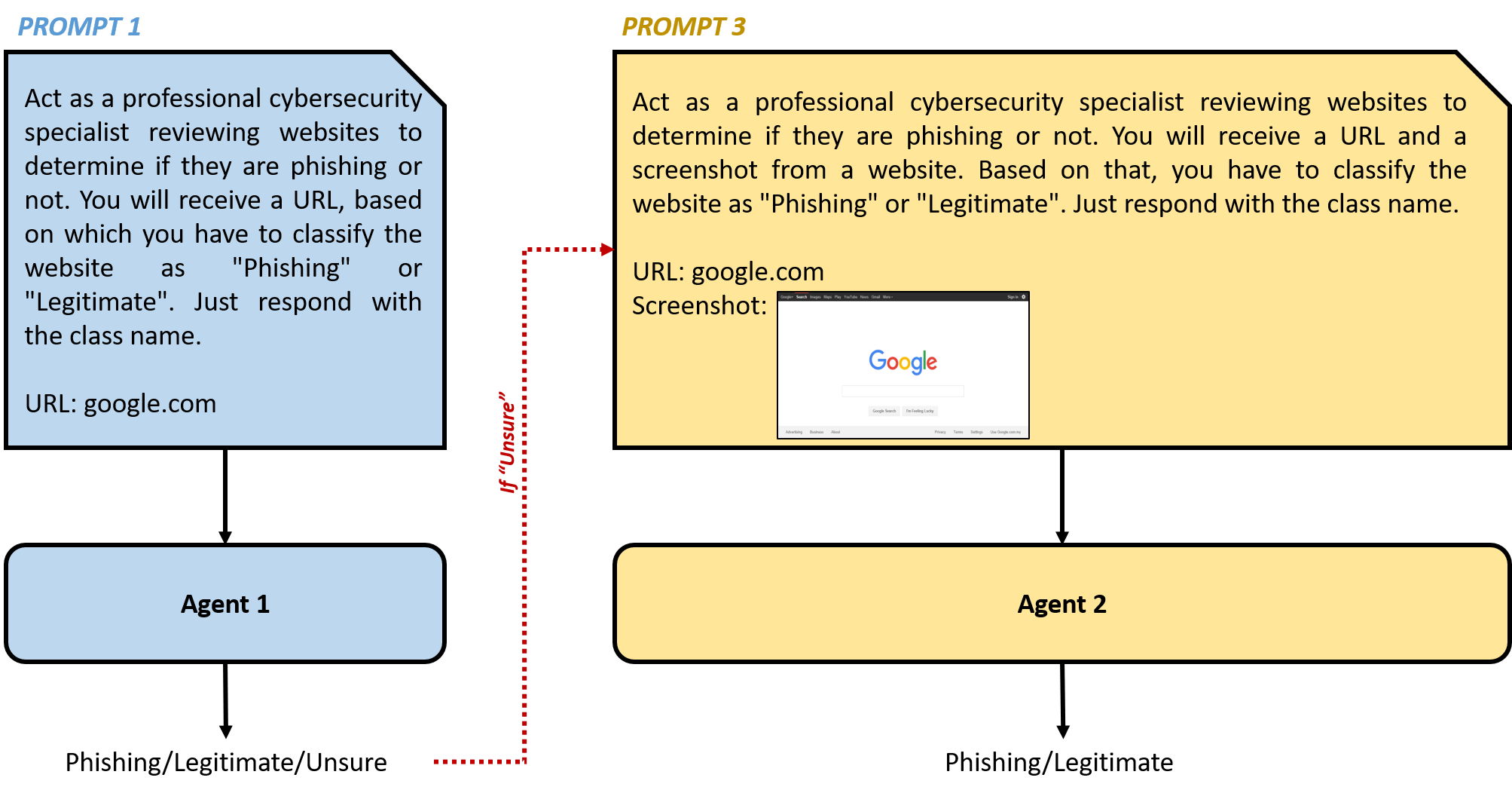}}
\caption{Agentic Approach: two Agents Operating Sequentially for Phishing Detection}
\label{fig:agentic}
\end{figure*}

The agentic approach involves a sequential decision-making process across two LMMs as shown in Figure \ref{fig:agentic}. The first LMM receives a prompt with only the URL and is asked to classify the website as phishing, legitimate, or unsure. If the first LMM confidently classifies the website (either as phishing or legitimate), the process ends. However, if the first response is uncertain, a second LMM is engaged. The second LMM receives a prompt that includes both the URL and a screenshot of the webpage for a more thorough analysis. This staged approach reduces the need to process both modalities for every input,  thereby lowering overall API costs.

\section{Experiments}
In this section, we describe the experimental setup used to evaluate the performance of the proposed phishing detection approaches. We detail the datasets, models, and evaluation metrics used in the experiments, followed by the results obtained from each approach.

\subsection{Experimental Setup}
\subsubsection{Data}
The experiments were conducted using a dataset provided by Chiew et al., which contains both phishing and legitimate websites \cite{chiew2018building}. The dataset comprises 30,000 websites, evenly split between 15,000 phishing and 15,000 legitimate sites, each including a URL, screenshot, HTML code, and additional information. We selected a stratified subset of 1,000 websites, evenly distributed between 500 phishing and 500 legitimate sites, to maintain the original dataset distribution while minimizing the costs associated with experiments involving multiple models and approaches, ensuring a representative sample set.

\subsubsection{Models}
For this study, we employed two LMMs, Gemini 1.5 Flash and GPT-4o mini, both capable of processing text-based and visual inputs. These models were chosen due to their accessibility via API and their relatively lower pricing compared to other LMMs, making them particularly suitable for industry settings focused on minimizing phishing threats while keeping operational costs manageable.

\subsubsection{Evaluation Metrics}
The effectiveness of each approach was measured using the following standard evaluation metrics:
\begin{enumerate}
    \item \textbf{Accuracy}: The proportion of correctly classified websites (both phishing and legitimate) out of the total number of websites tested.
    \item \textbf{Precision}: The proportion of websites classified as phishing that were actually phishing.
    \item \textbf{Recall}: The proportion of actual phishing websites correctly classified as phishing.
    \item \textbf{F1-Score}: The harmonic mean of precision and recall, providing a balanced measure of the model’s performance.
\end{enumerate}
 
\subsection{Results}
\begin{figure*}[htbp]
    \centering
    \begin{subfigure}{0.48\textwidth}
        \centering
        \includegraphics[width=\textwidth]{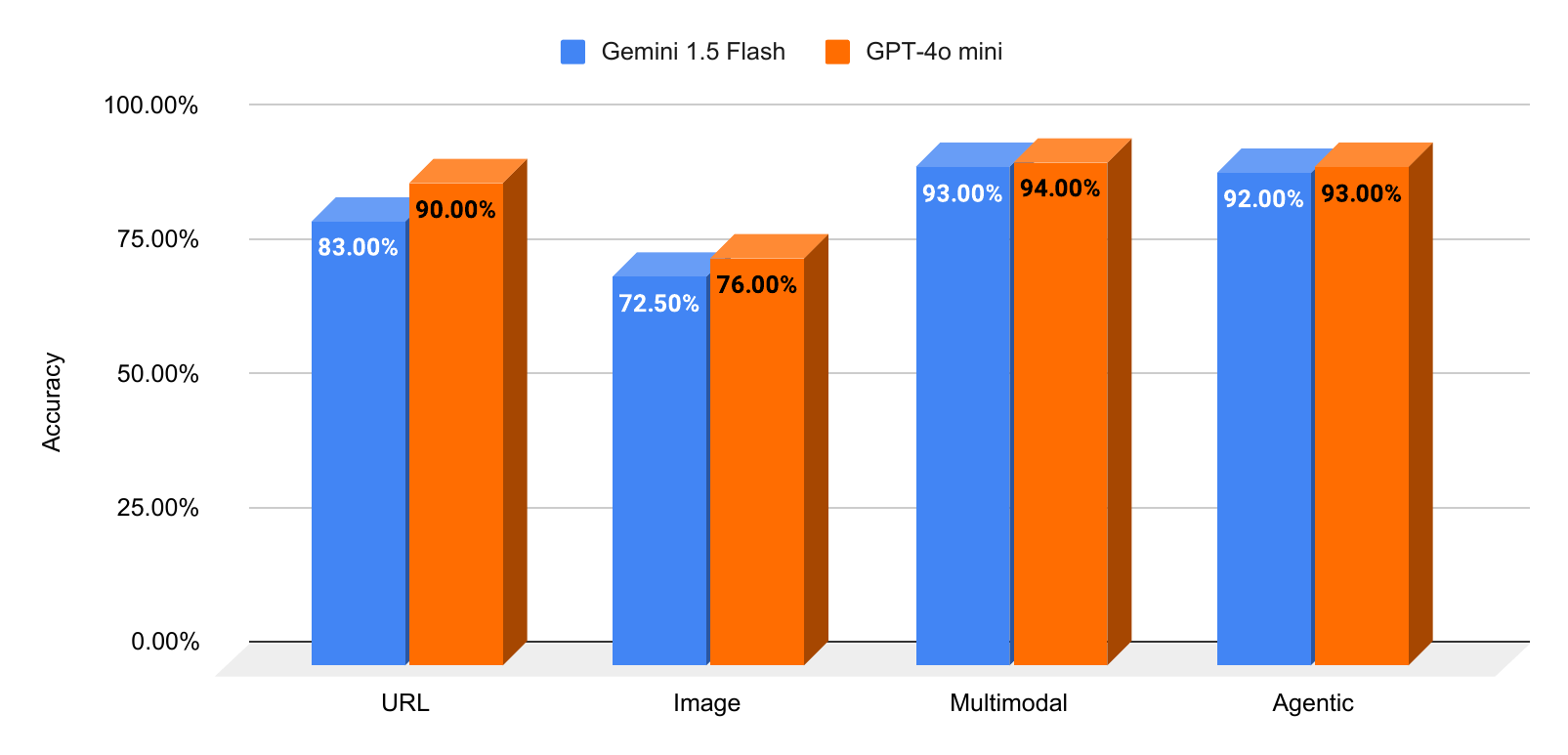}
        \caption{Accuracy}
        \label{fig:accuracy}
    \end{subfigure}
    \begin{subfigure}{0.48\textwidth}
        \centering
        \includegraphics[width=\textwidth]{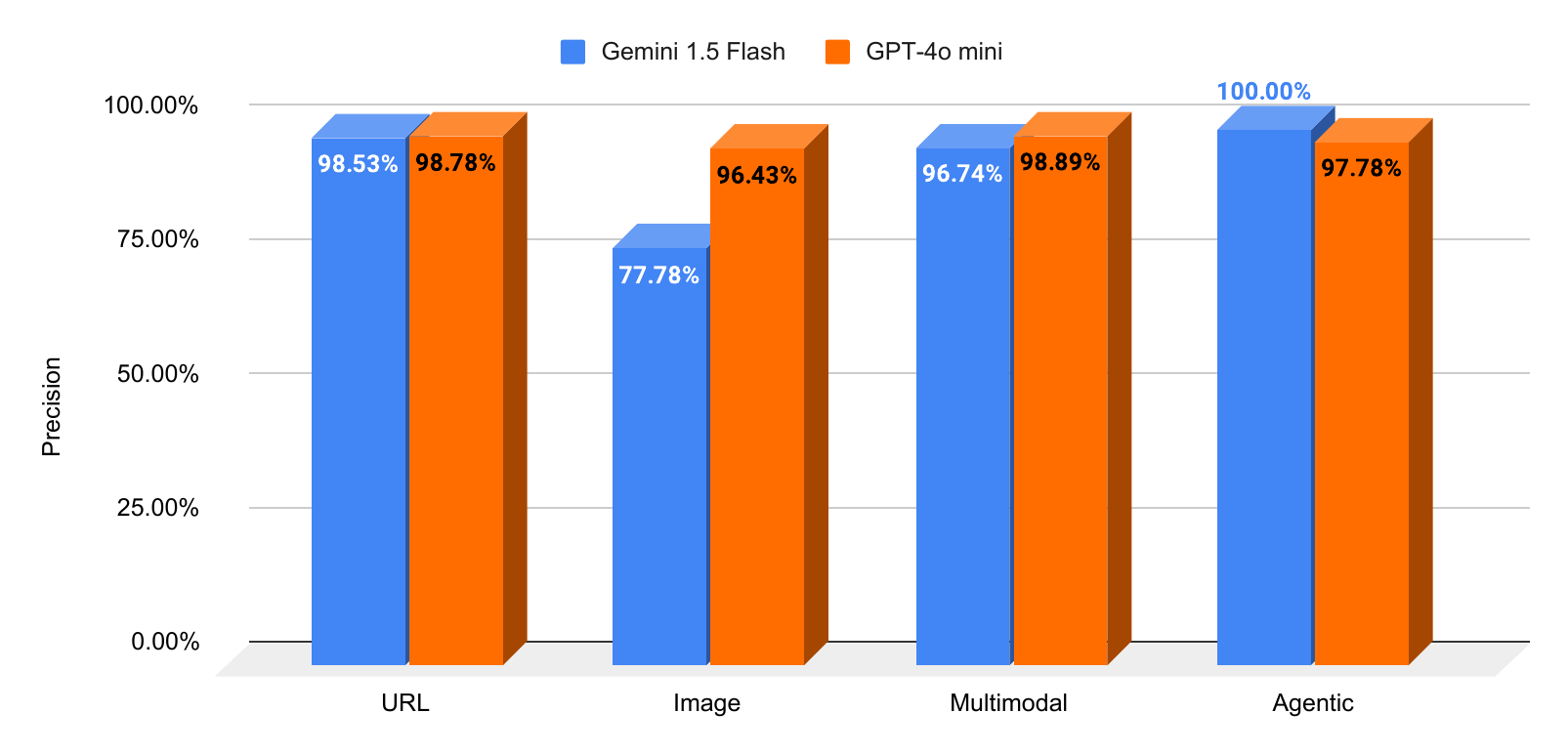}
        \caption{Precision}
        \label{fig:precision}
    \end{subfigure}
    
    \begin{subfigure}{0.48\textwidth}
        \centering
        \includegraphics[width=\textwidth]{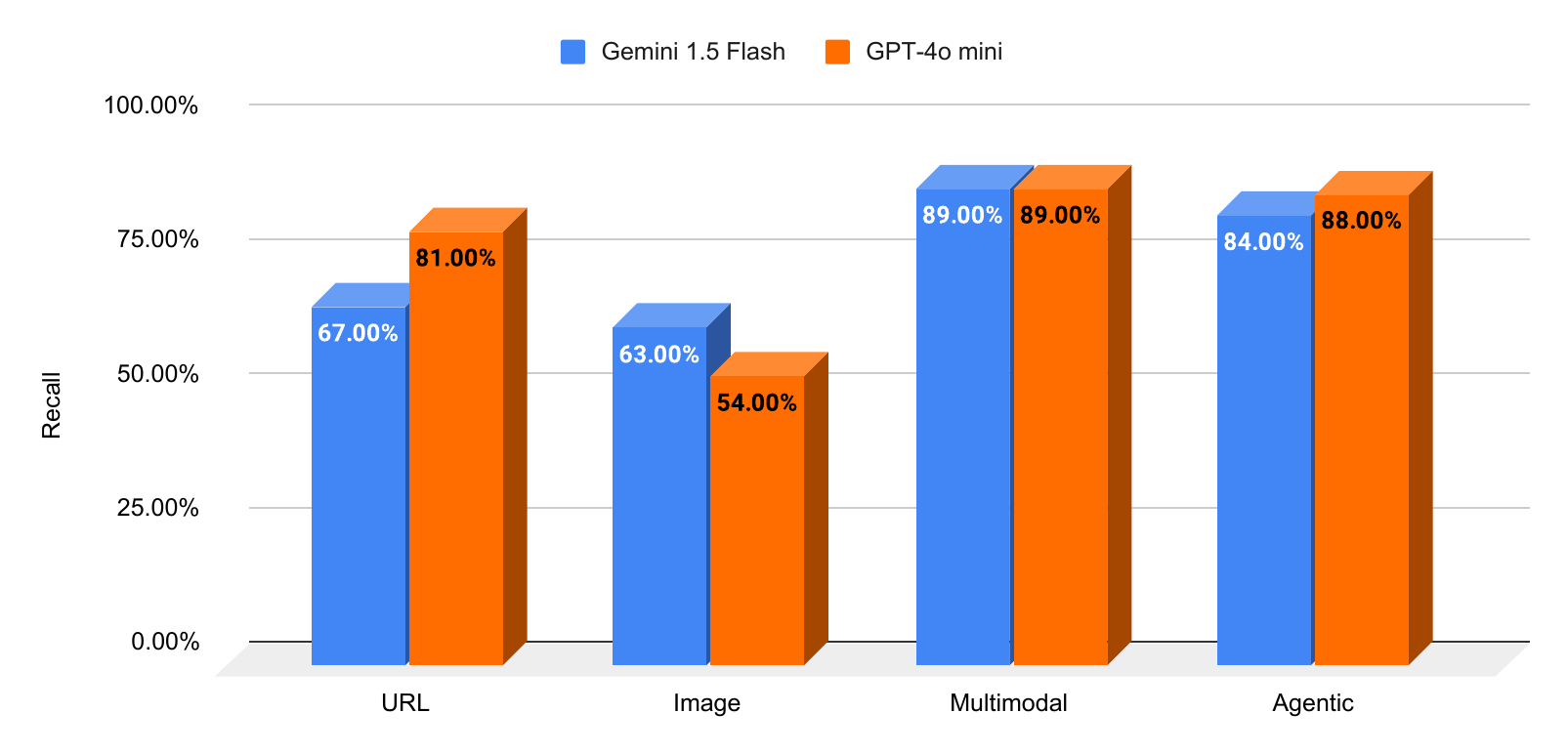}
        \caption{Recall}
        \label{fig:recall}
    \end{subfigure}
    \begin{subfigure}{0.48\textwidth}
        \centering
        \includegraphics[width=\textwidth]{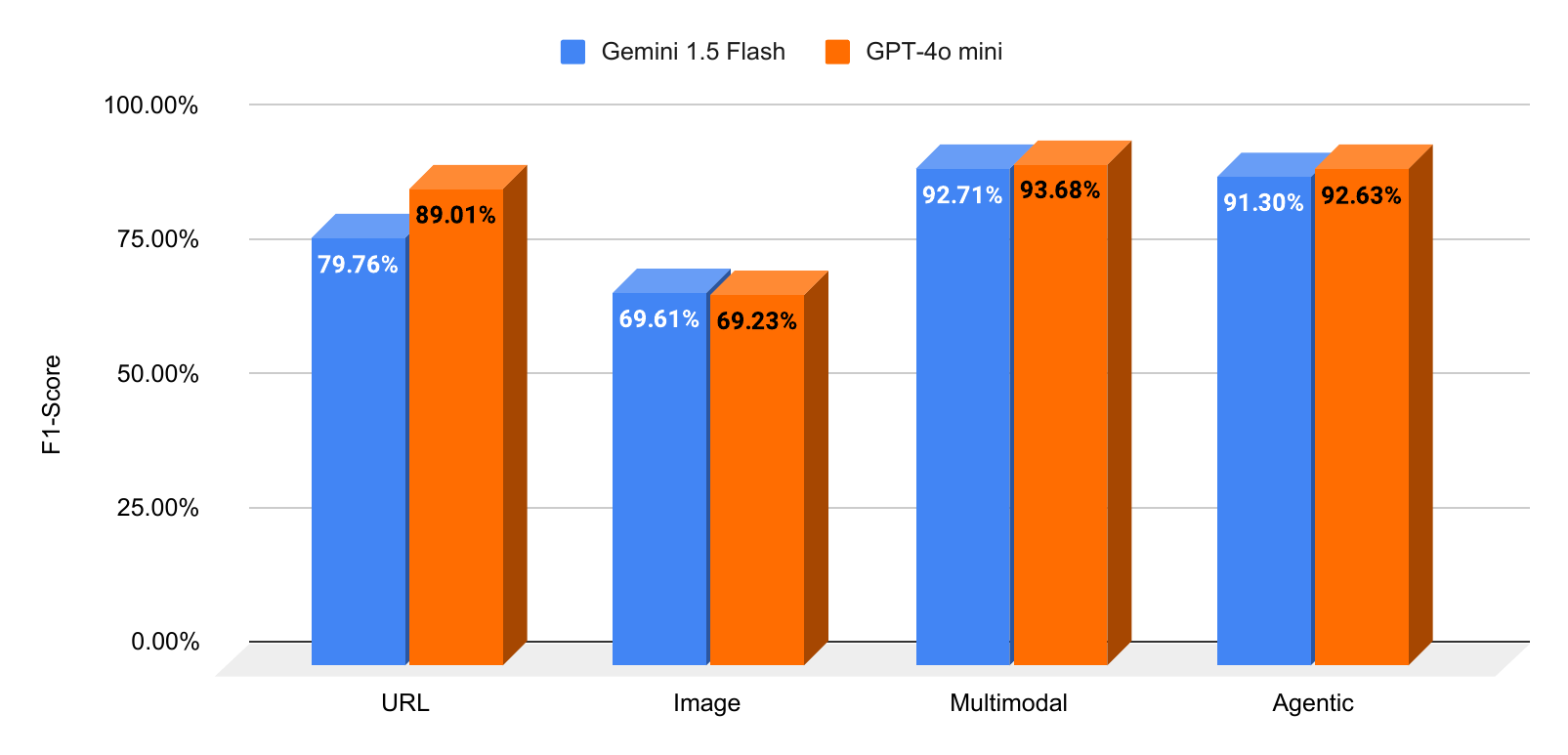}
        \caption{F1-Score}
        \label{fig:f1}
    \end{subfigure}

    \caption{Comparison of performance metrics across the four approaches for both models}
    \label{fig:performance-metrics}
\end{figure*}

The results obtained from applying the methodology to the dataset at hand are shown in Figure \ref{fig:performance-metrics}.
\subsubsection{URL-based Phishing Detection}
In the URL-based approach, the LMMs were prompted with URL strings and tasked with classifying websites as either phishing or legitimate. The models demonstrated varying levels of effectiveness. For the Gemini 1.5 Flash model, the approach achieved an accuracy of 83\%, with a precision of 98.53\% and a recall of 67\%, leading to an F1-score of 79.76\%. On the other hand, the GPT-4o mini model showed better performance, achieving an accuracy of 90\%, with a precision of 98.78\%, a recall of 81\%, and an F1-score of 89.01\%.
\subsubsection{Image-based Phishing Detection}
In the image-based approach, the LMMs were prompted with screenshots of the websites and asked to classify them. The Gemini 1.5 Flash model achieved an accuracy of 72.5\%, with a precision of 77.78\% and a recall of 63\%, resulting in an F1-score of 69.61\%. The GPT-4o mini model performed slightly better, with an accuracy of 76\%, a precision of 96.43\%, but a recall of only 54\%, yielding an F1-score of 69.23\%.
\subsubsection{Multimodal Phishing Detection}
The multimodal approach, which combines both URL and image inputs, provided the highest overall performance. For the Gemini 1.5 Flash model, this method resulted in an accuracy of 93\%, with a precision of 96.74\%, a recall of 89\%, and an F1-score of 92.71\%. Similarly, the GPT-4o mini model achieved an accuracy of 94\%, with a precision of 98.89\%, a recall of 89\%, and an F1-score of 93.68\%. 
\subsubsection{Agentic Phishing Detection}
The agentic approach utilized a sequential decision-making process, starting with URL analysis and escalating to multimodal analysis only when necessary. For the Gemini 1.5 Flash model, this approach achieved an accuracy of 92\%, with a precision of 100\%, a recall of 84\%, and an F1-score of 91.30\%. For the GPT-4o mini model, it resulted in an accuracy of 93\%, with a precision of 97.78\%, a recall of 88\%, and an F1-score of 92.63\%. 

\section{Discussion}
In this section, we analyze the results from both performance and cost perspectives, critically examining the strengths and weaknesses of each approach. The discussion highlights the trade-offs between detection accuracy and operational efficiency, which are essential considerations for deploying phishing detection systems in real-world environments.

\subsection{Performance Analysis}
The results indicate that the \textbf{multimodal approach} provides the highest accuracy and F1-score across both models. For the Gemini 1.5 Flash model, the multimodal approach reached 93\% accuracy with an F1-score of 92.71\%, while the GPT-4o mini model achieved 94\% accuracy with an F1-score of 93.68\%. These results are expected, as the multimodal approach combines the strengths of both URL and visual analysis, allowing the models to capture a wider range of phishing signals. For instance, phishing websites often manipulate both the URL (e.g., by using misleading domain names) and visual elements (e.g., by mimicking legitimate branding). By leveraging both inputs, the multimodal approach is better equipped to detect sophisticated phishing attempts that would otherwise evade single-modality detection.

However, the enhanced performance of the multimodal approach comes at a significant API cost, as it requires processing both URL text and visual data, leading to increased token usage and higher operational expenses. Additionally, the slight gap in performance between the two models suggests that while GPT-4o mini slightly outperforms Gemini 1.5 Flash, both models effectively utilize the multimodal data, indicating that the approach is robust across different LMM architectures.

On the other hand, the \textbf{URL-based detection approach}, while more cost-effective, showed some performance limitations. The Gemini 1.5 Flash model achieved an accuracy of 83\% with an F1-score of 79.76\%, whereas GPT-4o mini reached 90\% accuracy with an F1-score of 89.01\%. The high precision observed in both models (over 98\% for both) suggests that when a URL is classified as phishing, it is highly likely to be correct. However, the comparatively lower recall, especially for Gemini 1.5 Flash (67\%), highlights the model's struggle to identify a significant portion of phishing websites. This indicates that while URL-based detection is effective in minimizing false positives, it risks overlooking many phishing threats, particularly those using URLs that appear legitimate but have malicious content. Thus, URL-based detection is more suited for scenarios where high precision is prioritized over comprehensive coverage.

The \textbf{image-based detection approach} underperformed relative to the others, with the Gemini 1.5 Flash model achieving 72.5\% accuracy and an F1-score of 69.61\%, while the GPT-4o mini model managed 76\% accuracy with an F1-score of 69.23\%. These results reflect the inherent limitations of visual analysis in isolation. In theory, although the models might be able to identify visual cues commonly associated with phishing, such as incorrect logos or unusual layouts, they are prone to being misled by legitimate sites that simply have poor design or by phishing sites that replicate legitimate visuals effectively. The low recall (63\% for Gemini 1.5 Flash and 54\% for GPT-4o mini) underscores this issue, as many phishing websites successfully evade detection due to their visually convincing design.

The \textbf{agentic approach} presents an intriguing compromise between performance and cost efficiency. With an accuracy of 92\% and an F1-score of 91.30\% for Gemini 1.5 Flash, and 93\% accuracy with an F1-score of 92.63\% for GPT-4o mini, this method closely matches the performance of the full multimodal approach. Notably, the precision remains high (100\% for Gemini 1.5 Flash and 97.78\% for GPT-4o mini), which indicates that when the model classifies a site as phishing, it does so with confidence. The trade-off, however, is a slightly reduced recall (84\% and 88\%, respectively), suggesting that some phishing websites are missed when relying initially on the URL-based assessment. Nevertheless, this loss is mitigated by the substantial reduction in API costs, which is further explored in the cost analysis section.

\subsection{Cost Analysis}
The cost analysis was conducted using the token consumption and pricing data provided for both the GPT-4o mini \cite{noauthor_pricing_nodate} and Gemini 1.5 Flash \cite{noauthor_gemini_nodate} models on August 6, 2024, as shown in Tables \ref{table:gpt4o-mini} and \ref{table:gemini-flash}, respectively. The analysis reveals significant differences in cost-effectiveness between the multimodal and agentic approaches, as well as notable distinctions between the two models based on their token pricing and how they process images into tokens. At the time of writing, the GPT-4o mini model charged \$0.15 per 1M input tokens and \$0.6 per 1M output tokens. The Gemini 1.5 Flash model charged \$0.35 per 1M input tokens and \$1.05 per 1M output tokens. For the calculations in tables \ref{table:gpt4o-mini} and \ref{table:gemini-flash}, it is essential to note that the agentic approach may involve two API calls (one from each agent), and both input and output tokens are accounted for, which is why we observe an average of 3 output tokens per request in the agentic method compared to 2 in the multimodal approach. Additionally, when calculating the average tokens, we round up to the nearest whole token due to the fractional tokens generated in edge cases. This rounding is done to ensure cost estimates are conservative and avoid underestimating potential expenses, especially at scale.

\begin{table*}[htbp]
\caption{Cost and Token Analysis for GPT-4o-mini}
\centering
\begin{tabular}{|l|c|c|c|c|}
\hline
\multirow{2}{*}{} & \multicolumn{2}{c|}{\textbf{Multi-modal approach}} & \multicolumn{2}{c|}{\textbf{Agentic approach}} \\ \cline{2-5} 
 & \textbf{Input tokens} & \textbf{Output tokens} & \textbf{Input tokens} & \textbf{Output tokens} \\ \hline
\textbf{Total for 1,000 websites} & 26,006,995 & 2,000 & 6,192,395 & 2,490 \\ \hline
\textbf{AVG per website} & 26,007 & 2 & 6,193 & 3 \\ \hline
\multirow{2}{*}{\textbf{Price per website}}& \multicolumn{1}{c|}{\$0.003901} & \multicolumn{1}{c|}{\$0.000001} & \multicolumn{1}{c|}{\$0.000929} & \multicolumn{1}{c|}{\$0.000002} \\ \cline{2-5} 
 & \multicolumn{2}{c|}{\textbf{Total = \$0.003902}} & \multicolumn{2}{c|}{\textbf{Total = \$0.000931}} \\ \hline
\textbf{Number of websites for \$100} & \multicolumn{2}{c|}{25,626} & \multicolumn{2}{c|}{107,440} \\ \hline
\textbf{Price for 1M website checks} & \multicolumn{2}{c|}{\$3,902.25} & \multicolumn{2}{c|}{\$930.75} \\ \hline
\end{tabular}
\label{table:gpt4o-mini}
\end{table*}

\begin{table*}[htbp]
\caption{Cost and Token Analysis for Gemini 1.5 Flash}
\centering
\begin{tabular}{|l|c|c|c|c|}
\hline
\multirow{2}{*}{} & \multicolumn{2}{c|}{\textbf{Multi-modal approach}} & \multicolumn{2}{c|}{\textbf{Agentic approach}} \\ \cline{2-5} 
 & \textbf{Input tokens} & \textbf{Output tokens} & \textbf{Input tokens} & \textbf{Output tokens} \\ \hline
\textbf{Total for 1,000 websites} & 326,465 & 2,000 & 118,895 & 2,320 \\ \hline
\textbf{AVG per website} & 327 & 2 & 119 & 3 \\ \hline
\multirow{2}{*}{\textbf{Price per website}} & \multicolumn{1}{c|}{\$0.000114} & \multicolumn{1}{c|}{\$0.000002} & \multicolumn{1}{c|}{\$0.000042} & \multicolumn{1}{c|}{\$0.000003} \\ \cline{2-5} 
 & \multicolumn{2}{c|}{\textbf{Total = \$0.000116}} & \multicolumn{2}{c|}{\textbf{Total = \$0.000045}} \\ \hline
\textbf{Number of websites for \$100} & \multicolumn{2}{c|}{862,068} & \multicolumn{2}{c|}{2,232,142} \\ \hline
\textbf{Price for 1M website checks} & \multicolumn{2}{c|}{\$116} & \multicolumn{2}{c|}{\$44.80} \\ \hline
\end{tabular}
\label{table:gemini-flash}
\end{table*}

For the GPT-4o mini model, the multimodal approach, which processes both URL and image data for every website, consumed an average of 26,007 input tokens per website, resulting in a cost of \$0.003902 per website. Given a budget of \$100, this method would allow the processing of approximately 25,626 websites. In contrast, the agentic approach, which initially processes only the URL and escalates to multimodal analysis only when necessary, consumed an average of 6,193 input tokens per website, resulting in a cost of \$0.000931 per website. With the same \$100 budget, the agentic approach could process around 107,440 websites—more than four times as many as the multimodal approach. This significant cost reduction highlights the efficiency of the agentic method, especially when large volumes of websites need to be processed.

For the Gemini 1.5 Flash model, the multimodal approach consumed an average of 327 input tokens per website, resulting in a total cost of \$0.000116 per website. With a \$100 budget, this approach could process approximately 862,068 websites. The agentic approach, on the other hand, consumed an average of 119 input tokens per website, resulting in a total cost of \$0.000045 per website. This allows for the processing of roughly 2,232,142 websites with the same \$100 budget—more than double what the multimodal approach could handle.

The difference in costs between the two models is influenced by how they process images into tokens. The Gemini 1.5 Flash model, under default settings, resizes images into a fixed resolution before tokenization, resulting in a smaller, consistent token count per image. This fixed-size approach keeps token usage relatively low, making it more cost-effective when handling large batches of websites, especially in multimodal settings. In contrast, GPT-4o mini processes images according to their actual size using default tokenization settings. This leads to more variability and, generally, higher token consumption per image, which drives up the overall cost in multimodal scenarios.

When considering the cost for large-scale operations, such as processing 1 million website checks, the difference becomes even more apparent. For GPT-4o mini, the multimodal approach would cost approximately \$3,902.25, while the agentic approach reduces that to \$930.75. For Gemini 1.5 Flash, the multimodal approach costs \$116, while the agentic approach costs only \$44.80. These figures underscore the substantial cost benefits of using the agentic approach, especially when deploying systems at scale.

In summary, while the multimodal approach provides the best detection performance, the agentic approach offers a near-equivalent level of accuracy with dramatically lower costs. This pattern was clearly shown for both LMMs, despite their notable differences in pricing and performance. With its selective application of full multimodal analysis, this strategy finds an ideal balance between cost and performance, making it well-suited for real-world deployments where both scalability and accuracy are critical.

\subsection{Comparing to Previous Works}
While the literature contains numerous studies on phishing detection, few have explored the application of LLMs for this purpose. The differences in datasets and models used across these studies naturally lead to variations in performance metrics, making direct comparisons challenging. Nevertheless, we aim to compare the approaches and highlight their respective benefits.

Saha Roy et al. employed LLMs for phishing website detection by analyzing HTML page content, achieving impressive results with 97\% accuracy and a 96\% F1-score using LLaMA2 \cite{roy2024utilizing}. However, their approach relied heavily on fine-tuning LLMs, which introduces challenges related to ongoing maintenance and retraining to adapt to evolving phishing techniques.

Li et al. investigated the use of LLMs with zero-shot and few-shot prompts to detect phishing websites based on URL and textual content, achieving an F1-score of 90.2\% \cite{li2023prompting}. While their study focused on text data, our research incorporates image content as well, demonstrating the potential for improved detection performance.

Trad and Chehab compared prompt-engineered LLMs with fine-tuned LLMs, applying various prompt engineering strategies, including the role-playing prompt used in our study. Their results, specifically for the role-playing prompts, reported an accuracy of 85.19\% and 92.70\%, and an F1-score of 82.95\% and 92.26\% for GPT-3.5-Turbo and Claude 2, respectively \cite{trad2024prompt}. These results are comparable to our URL-based analysis, and the difference is attributed to the difference in datasets and models. However, the proposed multimodal and agentic approaches surpass these outcomes by incorporating both visual and textual data.

Wang et al. introduced an LLM-based agent framework for phishing detection that achieved an accuracy of 94.5\% \cite{wang2024automated}. Their method involved a single agent capable of performing function calls to analyze interface elements, URLs, and webpage text. Despite the sophistication of their approach, they tested it on only 400 samples due to the computational costs associated with analyzing complex HTML content. Additionally, their model's runtime was approximately 20 seconds per website, compared to the significantly faster processing times of 0.8 seconds for our agentic approach and 1.8 seconds for our multimodal approach. This highlights the efficiency and scalability advantages of our methods.

In summary, while prior work has made significant contributions to phishing detection using LLMs, our study offers the advantage of achieving strong performance by incorporating multiple modalities without additional training, while also reducing API costs and runtime.

\section{Conclusion}
This study explored multimodal AI agents for phishing detection. Our findings reveal that the multimodal approach offers the highest accuracy, leveraging both URL and visual inputs to identify phishing websites with superior precision and recall. However, this accuracy comes at the cost of significant API costs, especially when scaled to large volumes of websites.

The agentic approach emerged as a strong alternative, delivering a performance that closely matches the multimodal approach while dramatically reducing token consumption and overall costs. By employing a staged decision-making process, the agentic approach effectively balances detection accuracy with operational efficiency, making it highly suitable for real-world applications where both scalability and budget constraints are critical factors. 

Although the current study provides significant insights into the trade-offs between detection accuracy and cost efficiency, several avenues remain open for future research and improvement. Future work could explore incorporating multiple LLMs/LMMs within the agentic framework, where models with different strengths are strategically combined to balance cost and accuracy. For instance, a lightweight LLM could be used for initial URL screening, followed by a more advanced LMM for multimodal analysis only when needed, reducing overall token consumption. Additionally, optimizing prompt engineering to improve model certainty could further decrease the number of escalations to the more resource-intensive multimodal analysis. Another promising direction is integrating model ensembles, where multiple models collaborate by focusing on specific features (e.g., some models prioritizing text analysis, others focusing on visual content), potentially enhancing both accuracy and efficiency. Finally, exploring the real-time deployment of these agentic systems and evaluating their performance under varied phishing scenarios could provide deeper insights into their practical applications and adaptability in real-life scenarios.

\section*{Acknowledgment}
The authors would like to acknowledge that this work has been supported by the Maroun Semaan Faculty of Engineering and Architecture at the American University of Beirut. 

\bibliographystyle{ieeetr}
\bibliography{biblio}

\end{document}